\documentclass{article}

\usepackage[final,nonatbib]{nips_2016}

\usepackage[utf8]{inputenc} 
\usepackage[T1]{fontenc}    
\usepackage{url}            
\usepackage{booktabs}       
\usepackage{amsfonts}       
\usepackage{nicefrac}       
\usepackage{microtype}      
\usepackage{caption}        
\usepackage{subcaption}     

\usepackage{amsmath}
\usepackage{bbding}
\usepackage{graphicx}
\usepackage{xspace}

\usepackage{amsfonts}
\usepackage{xcolor}
\definecolor{darkgreen}{rgb}{0,0.6,0}
\definecolor{darkred}{rgb}{0.6,0,0}
\definecolor{darkblue}{rgb}{0,0.05,0.35}

\newcommand{\glb}{$\mathcal{G}_{\text{LB}}$}
\newcommand{\galt}{$\mathcal{G}_{\text{ALT}}$}

\title{Improved generator objectives for GANs}

\author{Ben Poole\thanks{Work done during an internship at Google Brain.}\\
Stanford University\\
\texttt{poole@cs.stanford.edu}
\And
Alexander A. Alemi, Jascha Sohl-Dickstein, Anelia Angelova\\
Google Brain\\
\texttt{\{alemi, jaschasd, anelia\}@google.com}
}

\begin{document}

\maketitle

\begin{abstract}
We present a framework to understand GAN training as alternating density ratio estimation, and approximate divergence minimization. This provides an interpretation for the mismatched GAN generator and discriminator objectives often used in practice, and explains the problem of poor sample diversity. Further, we derive a family of generator objectives that target arbitrary $f$-divergences without minimizing a lower bound, and use them to train generative image models that target either improved sample quality or greater sample diversity.
\end{abstract}

\section{Introduction}
Generative adversarial networks (GANs) have become a popular method for fitting latent-variable directed generative models to complex datasets \cite{Goodfellow14,Odena2016b,Salimans16, Ledig2016}. While these models provide compelling visual samples, they are notoriously unstable and difficult to train and evaluate. Many recent papers have focused on new architectures and regularization techniques for improved stability and performance \cite{Salimans16, Radford15, Huszar15}, but the objectives they optimize are fundamentally the same as the objectives in the original proposal \cite{Goodfellow14}.

The visual quality of samples from generative models trained with GANs often exceeds those of their variationally-trained counterparts \cite{Kingma13, Rezende2014}. This is often credited to a difference in the divergence between the data and model distribution that each technique optimizes \cite{Theis2016a}. GAN theory shows that an idealized formulation optimizes Jensen-Shannon divergence, while VAEs optimize a lower bound on log-likelihood, corresponding to a lower bound on the KL divergence. Recent work has generalized the GAN theory to target reverse KL \cite{Sonderby16} and additional $f$-divergences (including KL, reverse KL, and JS), allowing GANs to target a diverse set of behaviors \cite{Nowozin16}.

However, these new theoretical advances fail to provide a justification for the GAN objectives that are used in practice. In particular, the generator objective used in practice is different from the one that is theoretically justified \cite{Goodfellow14,Nowozin16}. This raises the question as to whether the theory used to motivate GANs applies to these modified objectives, and how the use of mismatched generator and discriminator objectives influences the behavior of GANs in practice.

Here we present a new interpretation of GANs as alternating between steps of density ratio estimation, and divergence minimization. This leads to a new understanding of the GAN generator objective that is used in practice as targeting a mode-seeking divergence that resembles reverse KL, thus providing an explanation for the mode dropping seen in practice. Furthermore, we introduce a set of new objectives for training the generator of a GAN that can trade off between sample quality and sample diversity, and show their effectiveness on CIFAR-10.

\section{Theory}
\subsection{Background}
Given samples from a data density, $x \sim q(x)$, we would like to learn a generative model with density $p$ that matches the data density $q$. Often the models we are interested in have intractable likelihoods, so that we can sample $x$ efficiently but cannot evaluate its likelihood.
In the GAN framework \cite{Goodfellow14}, the intractable likelihood is bypassed by instead training a discriminator to classify between samples from the data and samples from the model. Given this discriminator, the parameters of the generative model are updated to increase the tendency of the discriminator to mis-classify samples from the model as samples from the data. This iterative process pushes the model density towards the data density without ever explicitly computing the likelihood of a sample. More formally, the GAN training process is typically motivated as solving a minimax optimization problem:
\begin{equation}
\underset{p}{\text{minimize}} \; \underset{d}{\text{max}}\, \left(\mathbb{E}_{x\sim q} \left[\log d(x)\right] + \mathbb{E}_{x\sim p} \left[\log \left(1-d(x)\right)\right]\right)
\end{equation}
where $p$ is the generative model distribution, $d$ is the discriminator, and $q$ is the data distribution. Fixing $p$, the optimal discriminator is $d^*(x)=\frac{q(x)}{q(x)+p(x)}$ \cite{Goodfellow14}. Thus if the inner maximization over the discriminator is performed to completion for each step of $p$, the GAN objective is equivalent to minimizing: 
\begin{equation}
\underset{p}{\text{minimize}} \; \left(\mathbb{E}_{x\sim q}\left[\log \frac{q(x)}{q(x) + p(x)}\right] + \mathbb{E}_{x\sim p}\left[\log \left(1-\frac{q(x)}{q(x) + p(x)}\right)\right]\right)  = 2\,\text{JS}(q \| p) - \log 4
\end{equation}
This has led to the understanding that GANs minimize the Jensen-Shannon divergence between the data density and the model density, and is thought to underlie the difference in sample quality between GANs and VAEs \cite{Theis2016a}. However, this is not the objective that is used in practice, and we will see below that this alters the analysis.

Recently, \cite{Nowozin16} proposed an extension to GANs to target divergences other than Jensen-Shannon. They generalize the set of divergences a GAN can target to the family of $f$-divergences, where:
\begin{equation}
    D_f\left(q \| p\right) = \int dx\, p(x) f\left(\frac{q(x)}{p(x)} \right)
\end{equation}
and $f(u): \mathbb{R}^+ \to \mathbb{R}$ is a convex function with $f(1)=0$. The key result they leverage from \cite{nguyen2007} is that any $f$-divergence can be lower-bounded by 
\begin{equation}
D_f(q\| p) \ge \sup_{T\in \mathcal{T}} \left(\mathbb{E}_{x\sim q}\left[T(x)\right] - \mathbb{E}_{x\sim p}\left[f^\star(T(x))\right] \right)
\label{eq:lb}
\end{equation}
where $f^\star$ is the Fenchel conjugate\footnote{The Fenchel conjugate is defined as $f^\star(t) = \sup_{u \in \text{dom}_f} \left(ut - f(u)\right)$}
of $f$, and $T$ is the variational function also known as the discriminator in the GAN literature\footnote{We use $q$ as the data distribution and $p$ as the model distribution, which is the opposite of \cite{Nowozin16}.}. Thus for any $T$, we have a lower bound on the divergence that recovers exactly the discriminator objective used in the standard GAN when $f(u)=u\log u - (u+1)\log(u+1)$. As this is a lower bound on the $f$-divergence, maximizing it with respect to the discriminator $T$ makes sense, and yields a tighter lower bound on the true divergence.

However, the objective to optimize for the {\em generative model}, $p$, remains unclear. In both the original GAN paper \cite{Goodfellow14} and the $f$-GAN paper \cite{Nowozin16}, two objectives are proposed (denoted as \glb{} and \galt):
\begin{enumerate}
    \item \glb: Minimize the {\em lower} bound in Equation \ref{eq:lb}. For standard GANs, this corresponds to minimizing the probability of the discriminator classifying a sample from the model as fake. 
    \item \galt: Optimize an alternative objective:
    \begin{equation}
    \underset{p}{\text{minimize}} \; 
    \mathbb{E}_{x\sim p}\left[-T(x)\right]
    \label{eq:practice}
    \end{equation}
    For standard GANs, this corresponds to maximizing the log probability of the discriminator classifying a sample from the model as real.
\end{enumerate}

The first approach minimizes a lower bound, and thus improvements in the objective can correspond to making $D_f(q\|p)$ smaller, or, more problematically, by making the lower bound on $D_f(q\|p)$ looser. In practice this leads to slower convergence, and thus the first objective is not widely used.

The second approach is empirically motivated in \cite{Goodfellow14, Nowozin16} as speeding up training, and theoretically motivated by the observation that $p=q$ remains a fixed point of the learning dynamics. However, the behavior of this generator objective when the generative model does not have the capacity to realize the data density remains unclear. This is the regime we care about as most generative models do not have the capacity to exactly model the data.

\subsection{Discriminator as a density ratio estimator}
To address the theoretical and practical issues we first present a simple relationship between the discriminator and an estimate of the density ratio. Given known data and model densities, the optimal discriminator with respect to an $f$-divergence, $f_D$, was derived in \cite{Nowozin16} as:
\begin{equation}
T^*(x) = f_D'\left(\frac{q(x)}{p(x)}\right)
\end{equation}
where $f_D'$ is the derivative of $f_D$.
If $f_D'$ is invertible, we can reverse the relationship, and use the discriminator to recover the ratio of the data density to the model density:
\begin{equation}
\frac{q(x)}{p(x)} = \left(f_D'\right)^{-1}\left(T^*(x)\right)\approx 
\left(f_D'\right)^{-1}\left(T(x)\right)
\end{equation}
In practice we don't have access to the optimal discriminator $T^*(x)$, and instead use the current discriminator $T(x)$ as an approximation.

\subsection{A new set of generator objectives}
Given access to an approximate density ratio $q(x)/p(x)$, we can now optimize any objective that depends only on samples from $q$ or $p$ and the value of the density ratio. Conveniently, $f$-divergences are a family of divergences that depend only on samples from one distribution and the density ratio! Given samples from $p$ and an estimate of the density ratio at each point,  we can compute an estimate of the $f$-divergence, $f_G$ between $p$ and $q$:
\begin{align}
D_{f_G}\left(p \| q\right) &= \mathbb{E}_{x \sim p} \left[ f_G\left(\frac{q(x)}{p(x)}\right)\right] \approx \mathbb{E}_{x \sim p} \left[f_G \left( \left(f_D'\right)^{-1}\left(T(x)\right)\right)\right] 
\equiv \mathcal{G}_{f_D, f_G}
\label{eq:ratio}
\end{align}
where $\mathcal{G}_{f_D, f_G}$ is the generator objective, $f_G$ is the $f$-divergence targeted for the generator, and $f_D$ the $f$-divergence targeted for the discriminator. $f_G$ and $f_D$ need not be the same $f$-divergence. For non-optimal discriminators, this objective will be a biased approximation of the $f$-divergence, but is not guaranteed to be either an upper or lower bound on $f_G$.

Our new algorithm for GAN training iterates the following steps:
\begin{enumerate}
\item Optimize the discriminator, $T$, to maximize a lower-bound on $D_{f_D}\left(q \| p \right)$ using Equation \ref{eq:lb}. 
\item Optimize the generator, $p$, to minimize $\mathcal{G}_{f_D, f_G}$, using the estimate of the density ratio from the current discriminator, $T$, in Equation \ref{eq:ratio}. 
\end{enumerate}
While the first step is identical to the standard $f$-GAN training algorithm, the second step comprises a new generator update that can be used to fit a generative model to the data while targeting any $f$-divergence. In practice, we alternate single steps of optimization on each minibatch of data.

\subsection{Related work}
Several recent papers have identified novel objectives for GAN generators. In \cite{Sonderby16}, they propose a generator objective corresponding to $f_G$ being reverse KL, and show that it improves performance on image super-resolution. \cite{goodfellow2014distinguishability} identifies the generator objective that corresponds to minimizing the KL divergence, but does not empirically evaluate this objective.

Concurrent with our work, two papers propose closely related GAN training algorithms. In \cite{ratio2016}, they directly estimate the density ratio by optimizing a different discriminator objective that corresponds to rewriting the discriminator in terms of the density ratio:
\begin{equation}
D_f(q\| p) \ge \sup_{r} \left(\mathbb{E}_{x\sim q}\left[f'(r(x))\right] - \mathbb{E}_{x\sim p}\left[f'\left(f^\star(T(x))\right)\right] \right)
\end{equation}
This approach requires learning a network that directly outputs the density ratio, which can be very small or very large and in practice the networks that parameterize the density ratio must be clipped \cite{ratio2016}. We found estimating a function of the density ratio to be more stable, in particular using the GAN discriminator objective the discriminator $T(x)$ estimates $\log \frac{q(x)}{q(x)+p(x)}$. However, there are likely ways of combining these approaches in the future to directly estimate stable functions of the density ratio independent of the discriminator divergence.

More generically, the training process can be thought of as two interacting systems: one that identifies a statistic of the model and data, and another that uses that statistic to make the model closer to the data.  \cite{implicit2016} discusses many approaches similar to the one presented here, but do not present experimental results.

\section{Interpreting the GAN generator objective used in practice, \galt}
We can use our new family of generator objectives to better understand \galt, the objective that is used in practice (Eq. \ref{eq:practice}). Given that $f_D$ is the standard GAN divergence, we can solve for the generator divergence, $f_G$, such that $\mathcal{G}_{\text{ALT}} = \mathcal{G}_{f_D, f_G}$, yielding:
\begin{equation}
f_G\left(u\right) = \log\left(1 + \frac{1}{u}\right)
\end{equation}
Thus minimizing \galt{} corresponds to minimizing an approximation of the $f_G$ divergence between the data density and the model density, not minimizing the Jensen-Shannon divergence.

\begin{figure}[h] \centering
\includegraphics[width=0.9\textwidth]{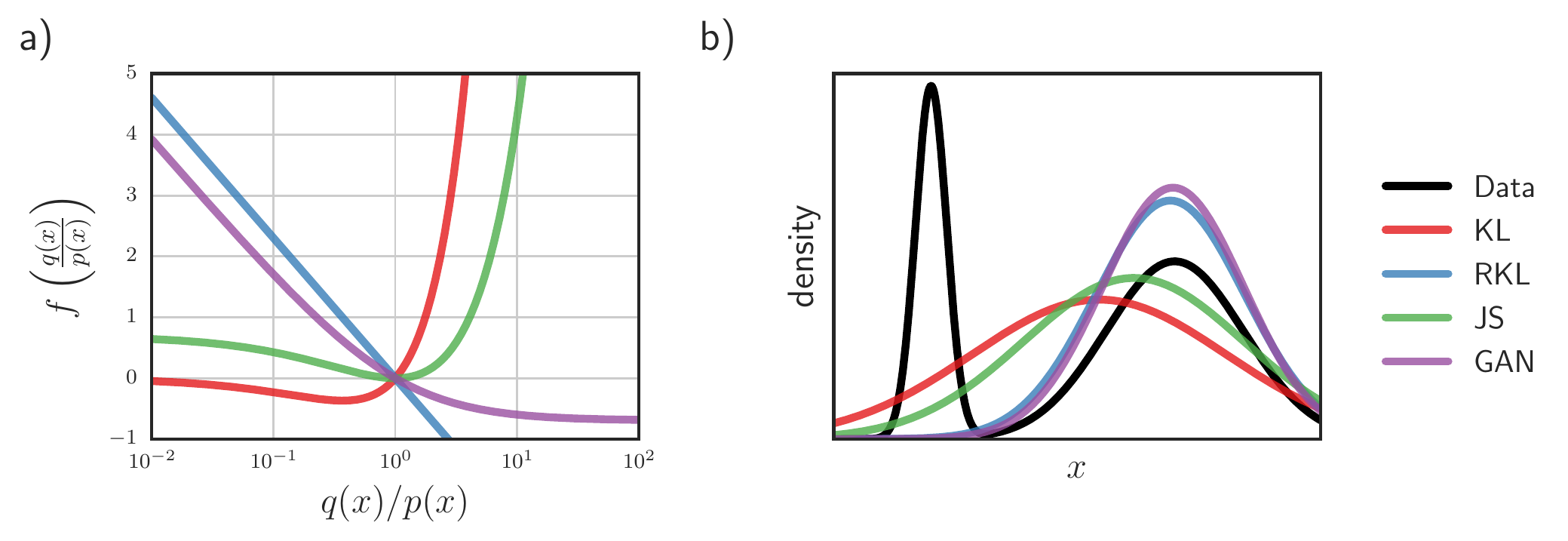}
\vspace{-4mm}
  \caption{The GAN generator objective used in practice (\galt) is mode-seeking when fit to a mixture of two Gaussians in one dimension. {\bf(a)} Value of the divergence function, $f$, as a function of the density ratio. The behavior of the GAN objective used in practice (\galt) resembles reverse KL when the model density is greater than the data density. 
  {\bf (b)} Learned densities when fitting a single Gaussian generative model to a mixture of two Gaussians (data, black). KL and JS are more mode-covering learning a generative model with larger variance that covers both modes of the data density, while reverse KL (RKL) and the GAN generator used in practice (GAN) are more mode-seeking, with smaller variance that covers only the higher density mode.
  }
 \label{fig:mog}
\end{figure}

To better understand the behavior of this divergence, we fit a single Gaussian to a mixture of two Gaussians in one dimension (Figure \ref{fig:mog}). We find that the GAN divergence optimized in practice is even more mode-seeking than JS and reverse KL. This behavior is likely the cause of many problems experienced with GANs in practice: samples often fail to cover the diversity of the dataset.

\begin{figure}[htbp]
  \centering
  \def \figuremult {0.44\textwidth}
  \begin{subfigure}[b]{\figuremult}
    \includegraphics[width=1.0\textwidth]{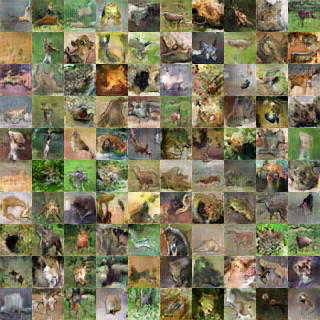}
    \caption{$\alpha=-3$}
    \label{fig:alpha3}
  \end{subfigure}
  \begin{subfigure}[b]{\figuremult}
    \includegraphics[width=1.0\textwidth]{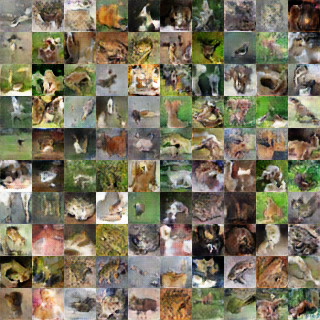}
    \caption{$\alpha=-1$}
    \label{fig:alpha1}
  \end{subfigure}
  \begin{subfigure}[b]{\figuremult}
    \includegraphics[width=1.0\textwidth]{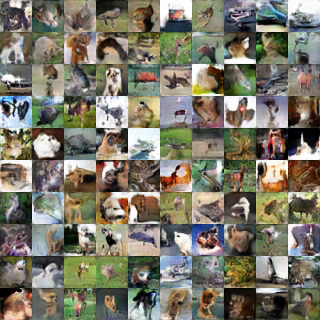}
    \caption{\galt{} (typical objective)}
    \label{fig:gan}
  \end{subfigure}
  \begin{subfigure}[b]{\figuremult}
    \includegraphics[width=1.0\textwidth]{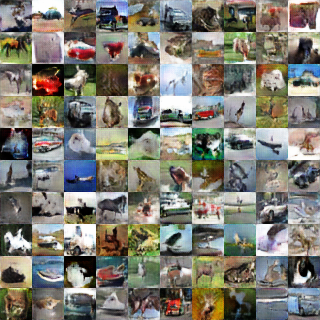}
    \caption{Reverse KL $\left( \alpha \to 0 \right)$}
    \label{fig:rkl}
  \end{subfigure}
  \begin{subfigure}[b]{\figuremult}
    \includegraphics[width=1.0\textwidth]{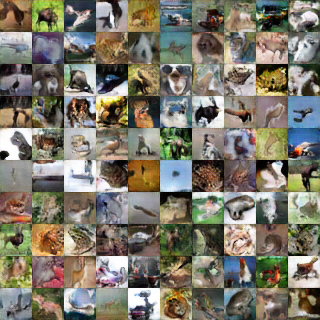}
    \caption{squared Hellinger $\left(\alpha=0.5\right)$}
    \label{fig:alpha0_5}
  \end{subfigure}
  \begin{subfigure}[b]{\figuremult}
    \includegraphics[width=1.0\textwidth]{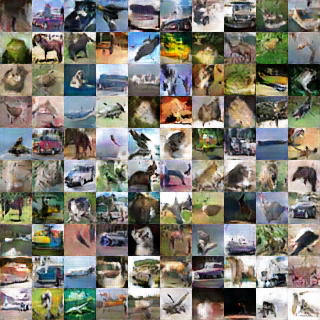}
    \caption{KL $\left(\alpha \to 1\right)$}
    \label{fig:kl}
  \end{subfigure}
  \caption{Different generator objectives yield different degrees of sample diversity. As we move from mode seeking $\alpha$-divergences with low $\alpha$ to mode covering divergences with $\alpha >0$ we see visual evidence of the increase in sample diversity, without a noticeable decrease in sample quality. In particular, note the overabundance of green and brown tones in the most mode seeking objectives. Sub-captions give the targeted generator divergence and are ordered from the most mode seeking to most mode covering. In all cases, the discriminator was trained using the standard GAN objective.}
 \label{fig:samples}
\end{figure}

\section{Experiments}
We evaluate our proposed generator objectives at improving the sample quality and diversity on CIFAR-10. All models were trained using identical architectures and hyperparameters (see Appendix B). The discriminator in all models was trained to optimize the normal GAN objective, corresponding to maximizing Equation \ref{eq:lb} with $f_D(u)=u\log u - (u+1)\log (u+1)$, and using $T(x) = g_f(V(x))$ with $V(x)\in \mathbb{R}$ being the output of a neural network and $g_f(v)=-\log(1+\exp(v))$ being used to constrain the range of $T$ as in \cite{Nowozin16}. For each model, we optimized a different generator objective by using different values for $f_G$ in Equation \ref{eq:ratio}. The generator objectives are derived and listed in Appendix A.

In order to highlight the effect the generator objective can have on the generated samples, we targeted several objectives at various $\alpha$ divergences, as well as the traditional generator objective \galt. 
In Figure \ref{fig:samples}, we see that the generator objective has a large impact on sample diversity. In particular, for very mode-seeking divergences ($\alpha=-3$ and $\alpha=-1$), the samples 
fail to capture the diversity of class labels in the dataset, as is immediately visually obvious from over-representation of greens and browns in the generated samples.
For more mode-covering divergences ($\alpha=0.5$ (squared Hellinger), KL) we see much better diversity in colors and sampled classes, without any noticeably degradation in sample quality.

\section{Discussion}
Our work presents a new interpretation of GAN training, and a new set of generator objectives for GANs that can be used to target any $f$-divergence. We demonstrate that targeting JS for the discriminator and targeting other objectives for the generator yields qualitatively different samples, with mode-seeking objectives producing less diverse samples, and mode-covering objectives producing more diverse samples.
However, training with very mode-seeking objectives does not yield extremely high-quality samples. Similarly, targeting mode-covering objectives like KL improves sample diversity, but the quality of samples does not visibly worsen. 
Visual evaluation of sample quality is a potentially fraught measure of quality however. 
Future work will be needed to investigate the impact of alternate generator objectives and provide better quantitative metrics and understanding of what factors drive sample quality and diversity in GANs.

\subsubsection*{Acknowledgments}
We thank Augustus Odena for feedback on the manuscript, Vincent Dumoulin for the baseline code, and Luke Metz, Luke Vilnis, and the Google Brain team for valuable and insightful discussions.
\bibliography{nips_2016}
\bibliographystyle{plain}

\appendix
\section{Deriving the generator objectives}
Here we derive the generator objectives when the discriminator divergence is $f_D(u) = u\log u - (u+1)\log (u+1)$, corresponding to the standard GAN discriminator objective. As in \cite{Nowozin16}, we parameterize the discriminator as $T(x) = g_f(V(x))$ where $g_f$ has the same range as $f_D'$. For the GAN case, this corresponds to $g_f(v)=-\log(1+\exp(-v))$.

First, we can compute the inverse of the gradient of $f_D$ which is used to estimate the density ratio: $$(f_D')^{-1}(t) = -\frac{e^t}{e^t-1}$$
For GANs, the discriminator is parameterized as $T(x) = -\log(1+\exp(-V(x))$, so we can compute the density ratio as:
\begin{equation}
\frac{q(x)}{p(x)} \approx -\frac{e^{T(x)}}{e^{T(x)}-1} = e^{V(x)}
\end{equation}

Given this estimate of the density ratio, we can then compute the generator objective as $f_G(e^{V(x)})$. The table below contains the generator objectives for many different $f_G$ given $f_D(u)=u\log u - (u+1)\log(u+1)$:

\begin{equation}
\begin{array}{|r|r|c|c|}
\hline
\text{Name} & \text{Generator \textit{f}-divergence } (f_G)  & \text{Generator objective (minimized)}\\
\hline
\text{GAN-standard} & \log(1 + \frac{1}{u}) & \log\left(1+e^{-V(x)}\right)=-T(x)\\
\text{GAN-RKL} & -\log u & -V(x)\\
\text{GAN-KL} & u\log u & V(x)e^{V(x)}\\
\text{GAN-}\alpha & \frac{1}{\alpha(\alpha-1)} \left(u^\alpha-1-\alpha(u-1)\right) & \frac{1}{\alpha(\alpha-1)} \left(e^{\alpha V(x)}-1-\alpha(e^{V(x)}-1)\right) \\
\hline
\end{array}
\end{equation}

\pagebreak
\section{CIFAR-10 architecture details}
This is a slightly modified version of the architecture from \cite{Dumoulin16}. Input images were scaled from $[0, 255]$ to $[0, 1]$.
\begin{table}[h]
\centering
\begin{tabular}{@{}rllllll@{}} \toprule
Operation              & Kernel       & Strides      & Feature maps & BN?      & Dropout & Nonlinearity \\ \midrule
$G_x(z)$ -- $64 \times 1 \times 1$ input                                                              \\
Transposed convolution & $4 \times 4$ & $1 \times 1$ & $256$        & $\surd$  & 0.0     & Leaky ReLU \\
Transposed convolution & $4 \times 4$ & $2 \times 2$ & $128$        & $\surd$  & 0.0     & Leaky ReLU \\
Transposed convolution & $4 \times 4$ & $1 \times 1$ & $64$         & $\surd$  & 0.0     & Leaky ReLU \\
Transposed convolution & $4 \times 4$ & $2 \times 2$ & $32$         & $\surd$  & 0.0     & Leaky ReLU \\
Transposed convolution & $5 \times 5$ & $1 \times 1$ & $32$         & $\surd$  & 0.0     & Leaky ReLU \\
Convolution            & $1 \times 1$ & $1 \times 1$ & $32$         & $\surd$  & 0.0     & Leaky ReLU \\
Convolution            & $1 \times 1$ & $1 \times 1$ & $3$          & $\times$ & 0.0     & Sigmoid    \\
$V(x)$ -- $3 \times 32 \times 32$ input                                                               \\
Convolution            & $5 \times 5$ & $1 \times 1$ & $32$         & $\times$ & 0.2     & Maxout     \\
Convolution            & $4 \times 4$ & $2 \times 2$ & $64$         & $\times$ & 0.5     & Maxout     \\
Convolution            & $4 \times 4$ & $1 \times 1$ & $128$        & $\times$ & 0.5     & Maxout     \\
Convolution            & $4 \times 4$ & $2 \times 2$ & $256$        & $\times$ & 0.5     & Maxout     \\
Convolution            & $4 \times 4$ & $1 \times 1$ & $512$        & $\times$ & 0.5     & Maxout     \\
Convolution            & $1 \times 1$ & $1 \times 1$ & $1024$       & $\times$ & 0.5     & Maxout     \\
Convolution            & $1 \times 1$ & $1 \times 1$ & $128$       & $\times$ & 0.5     & Maxout     \\
Convolution            & $1 \times 1$ & $1 \times 1$ & $1$          & $\times$ & 0.5     & Linear    \\ \midrule
Optimizer              & \multicolumn{6}{@{}l@{}}{Adam ($\alpha = 10^{-4}$, $\beta_1 = 0.5$, $\beta_2 = 0.999$)} \\
Batch size             & \multicolumn{6}{@{}l@{}}{128}												  \\
Leaky ReLU slope, maxout pieces       & \multicolumn{6}{@{}l@{}}{0.1, 2}                                                \\
Weight, bias initialization  & \multicolumn{6}{@{}l@{}}{Isotropic gaussian ($\mu = 0$, $\sigma = 0.01$), Constant($0$)} \\ \bottomrule
\end{tabular}
\vspace{0.2cm}
\caption{\label{tab:cifar10_description} CIFAR10 model hyperparameters. Maxout
    layers are used in the discriminator.}
\end{table}

\end{document}